\documentclass{sig-alternate}
\usepackage{mathptmx} % This is Times font

\usepackage{fancyhdr}
\usepackage[normalem]{ulem}
\usepackage[hyphens]{url}
\usepackage[sort,nocompress]{cite}
\usepackage[final]{microtype}
\usepackage[keeplastbox]{flushend}
\usepackage[bookmarks=true,breaklinks=true,letterpaper=true,colorlinks,citecolor=blue,linkcolor=blue,urlcolor=blue]{hyperref}
\usepackage{microtype}
\usepackage{graphicx}
\usepackage{subfigure}
\usepackage{booktabs} % for professional tables
\usepackage{amssymb}
\usepackage{threeparttable, multirow}
\fancypagestyle{firstpage}{
  \fancyhf{}
  \renewcommand{\headrulewidth}{0pt}
  \fancyfoot[C]{\thepage}
}

\usepackage{xspace}
\newcommand{\prjname}{eDKM\xspace}

\title{\prjname:  An Efficient and Accurate Train-time\\ Weight Clustering for Large Language Models} 
\begin{document}

\numberofauthors{8} %  in this sample file, there are a *total*
\author{
% 1st. author
\alignauthor
Minsik Cho \\
       \affaddr{Apple. USA}\\
       \email{minsik@apple.com}
\alignauthor
Keivan A. Vahid \\
       \affaddr{Apple. USA}\\
       \email{kalizadehvahid@apple.com}
\alignauthor
Qichen Fu \\
       \affaddr{Apple. USA}\\
       \email{qfu22@apple.com}
       \and
\alignauthor
Saurabh Adya \\
       \affaddr{Apple. USA}\\
       \email{sadya@apple.com}
\alignauthor
Carlo C Del Mundo \\
       \affaddr{Apple. USA}\\
       \email{cdelmundo@apple.com}
\alignauthor
Mohammad Rastegari \\
       \affaddr{Apple. USA}\\
       \email{mrastegari@apple.com}
       \and
\alignauthor       
Devang Naik \\
       \affaddr{Apple. USA}\\
       \email{naik.d@apple.com}\alignauthor
Peter Zatloukal \\
       \affaddr{Apple. USA}\\
       \email{pzatloukal@apple.com}
}

\maketitle
\thispagestyle{firstpage}
\pagestyle{plain}

%%%%%% -- PAPER CONTENT STARTS-- %%%%%%%%

\begin{abstract}
Since Large Language Models or LLMs have demonstrated high-quality performance on many complex language tasks, there is a great interest in bringing these LLMs to mobile devices for faster responses and better privacy protection. However, the   size of LLMs (i.e.,   billions of parameters) requires highly effective compression to fit into storage-limited devices. Among many compression techniques, weight-clustering, a form of non-linear quantization, is one of the leading candidates for LLM compression, and supported by modern smartphones. Yet, its training overhead is prohibitively significant for LLM fine-tuning. Especially, Differentiable KMeans Clustering, or DKM, has shown the state-of-the-art trade-off between compression ratio and accuracy regression, but its large memory complexity makes it nearly impossible to apply to train-time LLM compression. In this paper, we propose a memory-efficient DKM implementation, eDKM powered by novel techniques to reduce the memory footprint of DKM by orders of magnitudes. For a given tensor to be saved on CPU for the backward pass of DKM, we compressed the tensor by applying uniquification and sharding after checking if there is no duplicated tensor previously copied to CPU.
Our experimental results demonstrate that \prjname can
fine-tune and compress a pretrained LLaMA 7B model from 12.6 GB  to 2.5 GB (3bit/weight) with the Alpaca dataset by reducing
the train-time memory footprint of a decoder layer by 130$\times$, while delivering good accuracy on broader LLM benchmarks (i.e., 77.7\% for PIQA, 66.1\% for Winograde, and so on).
\end{abstract}
\section{Introduction}
Large language models or LLMs, and especially Generative Pre-trained Transformer  (GPT) models have shown excellent performance on many complex language tasks~\cite{ouyang2022training, zhang2022opt}. Such breakthrough leads to the desire  to run these LLMs locally on mobile devices  for user privacy~\cite{haq,pmlr-v80-wu18h}, but   even small   LLMs are too big for on-device execution. For example, the smallset LLaMA model has 7B parameters which is 14GB in FP16~\cite{touvron2023LLaMA}, while   high-end mobile devices have only up to 18GB DRAM.
Therefore, aggressively compressing LLMs via train-time optimizations, such as sparsification, quantization, or weight clustering, is a crucial step for on-device LLM deployment~\cite{haq, hawq-v2, Park_2018_ECCV,xor_net,quant_noise,stock2020bit, zhou2019neural, park2019lookahead, yu2018nisp, polino2018model, dkm,deepcomp_iclr16, stock2020bit, ullrich2017soft}

% ,pmlr-v80-wu18h, ullrich2017soft,stock2020bit,dkm}.

However, train-time optimization of LLM is highly expensive due to the model size and computational resource overheads.
Especially, the computational resource demand from  a train-time differentiable weight clustering in DKM~\cite{dkm}, one of the state-of-the-art weight clustering algorithm is prohibitively high, as
it needs to analyze the interactions between all the weights and all possible clustering options.
Accordingly, many existing LLM compression techniques, such as GTPQ~\cite{frantar2023gptq}, and AWQ~\cite{lin2023awq} rely on post-training optimization. 

In this work, we propose memory optimization techniques 
to enable train-time weight clustering and their applications to DKM~\cite{dkm}, leading to \prjname.
Our techniques include cross-device tensor marshaling and weight matrix uniquification/sharding.
When we used \prjname to fine-tune and compress LLaMA 7B model into 3bit-per-weight, we achieved  about 130$\times$ memory footprint reduction for a decoder stack, yet outperformed the existing 3bit compression techniques.

% \mohammad{be clear if it is train time compression and mention about accuracy and size trade off at inference}

% we revisit the idea of applying train-time weight clustering for LLM compression~\cite{coreml}.
% Weight clustering can offer a better trade-off between model accuracy and size than  quantization~\cite{deepcomp_iclr16,pmlr-v80-wu18h, ullrich2017soft,stock2020bit,dkm}, but at higher training cost, which accordingly motivated a body of research in post-training LLM compression~\cite{frantar2023gptq,lin2023awq}.

% In this work, we propose a 
% % has higher flexibility than train-time weight quantization~\cite{liu2023llmqat} and can deliver the state-of-the-art trade-off between model accuracy and size but exploiting such extra flexibility requires more system resources.

% % \mohammad{You should mention why no-one revisit it before ? what was the challenge? }
% More specifically, we propose \prjname for LLM compression, a memory-optimized version of DKM~\cite{coreml_dkm,dkm} (which is known to have very large memory complexity).

% \input{sections/related}

\begin{figure}[!t]
% \vspace{-0.1 in}

\centering 

\includegraphics[width=3in]{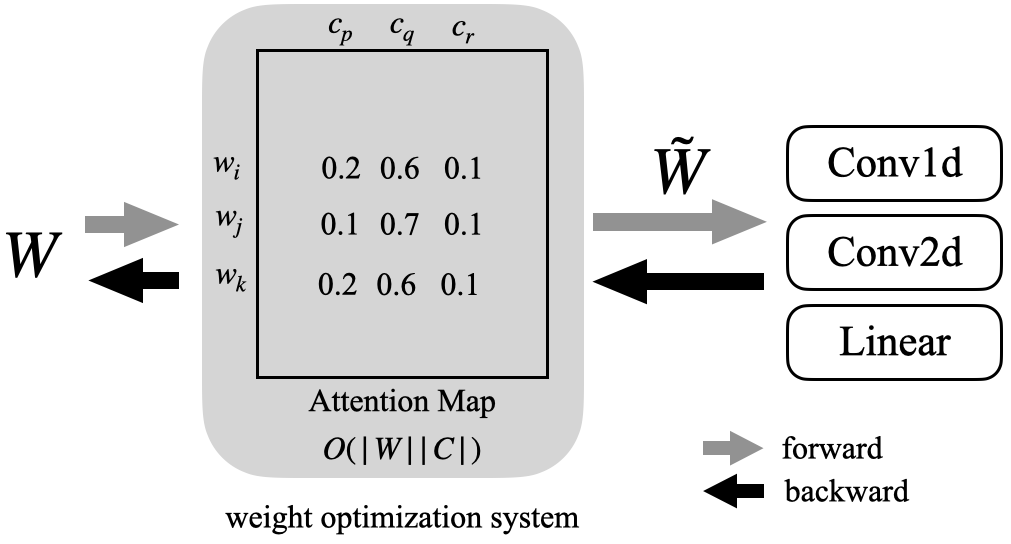}

 \caption{General overview of weight optimization systems. For DKM~\cite{dkm}, an attention map for differentiable weight clustering is created inside the system. }
 % \vspace{-0.15 in}
\label{mem_dkm:wopt_system}
\end{figure}

\section{Memory-efficient DKM}
% \mohammad{Overall you need to be clear this paper is about the training efficiency of DKM for LLMs(or in general) and as a result, we achieve state-of-the-art quantization-accuracy trade off in LLMs because DKM is the state-of-the-art quantization method}
Pruning, quantization, and normalization are all popular weight optimization techniques/systems that
take  in the original weights, $W$ and output  optimized weights $\tilde{W}$ for inference latency, test accuracy, or model size, as shown in Fig~\ref{mem_dkm:wopt_system}.
Among techniques, we focus on weight clustering, notably the state-of-the-art train-time weight clustering algorithm, DKM~\cite{dkm}. Weight clustering is a non-linear weight discretization, and a weight matrix will be compressed into a lookup table and a list of low-precision indices to the lookup table, which can be consumed by modern inference accelerators~\cite{coreml}.

DKM performs differentiable weight clustering by analyzing
 the interaction between the weights (denoted $W$) and centroids (denoted $C$), and has shown state-of-the-art  trades-off between compression ratio and accuracy. Therefore, using DKM for LLM compression would yield high-quality result.
 However, DKM computes a large  attention map with $O(|W||C|)$ memory complexity (i.e., the matrix in Fig.~\ref{mem_dkm:wopt_system}) for  forward/backward passes (see the Appendix in~\cite{dkm}), which is particularly challenging for LLM compression.
For example, a LLaMA 7B model needs at least 224GB just to compute an attention map for 4bit weight clustering.

% and   DKM~\cite{dkm} has shown state-of-the-art  trades off between compression ratio and accuracy by co-optimizing the weight gradient and the cluster centroids against the task loss. Therefore, 

Accordingly, we need to tap onto CPU memory to handle such large memory demand by overflowing to CPU memory and copying back to GPU when needed later. However, it will incur significant traffic between GPU and CPU (slowing down the training), and need immense CPU memory capacity.
Hence, it is critical to reduce the number of transactions between CPU and GPU, and minimize the traffic of each transaction. 
To address such challenges, we introduce two novel memory optimization techniques   in PyTorch.
 
\begin{itemize}  
   \item Cross-Device Tensor Marshaling: We track tensors being copied across devices and avoid redundant copying to reduce the memory footprint and expedite training.
   \item Weight Uniquification and Sharding: We use the fact that weights in 16 bits have only $2^{16}$ unique values to reduce the attention map (in Fig~\ref{mem_dkm:wopt_system}) representation and further shard it over multiple learners.
% \vspace{-0.3 in}    
\end{itemize}

\subsection{Cross-device Tensor Marshaling}
% \mohammad{it all describes what is the Cross-device Marshaling but it is not clear why it is important for DKM? not clear why in DKM you should move data across devices ? you need to first set the problem and then propose the solution}

PyTorch represents a tensor with data storage that links to the actual data layout and metadata that keeps
the tensor shapes, types, and so on.
Such tensor architecture lets PyTorch reuse the data storage whenever possible and efficiently reduces the memory footprint.
However, when a tensor moves to another device (i.e., from GPU to CPU), the data storage cannot be reused and a new tensor needs to be created. 
Table~\ref{cross_device_ex} shows an example of the memory footprint overhead when a tensor moves between devices in PyTorch. 
The tensor, $x_0$ allocated in line 0, consumes 4MB on GPU. When its view is changed in line 1, no additional GPU memory is required as the underlying data storage can be reused (i.e., $x_0$ and $x_1$ are effectively identical). However, when $x_0$ and $x_1$ move to CPU as in lines 2 and 3, the CPU memory consumption becomes 8MB, although $y_0$ and $y_1$ could share the same data storage on CPU, which leads to the redundancy on CPU memory and increases GPU-CPU traffic.

\begin{table}[!t]
\setlength{\tabcolsep}{8 pt}
\centering
\begin{threeparttable}

\begin{tabular}{c|c|cccccc|ccccc} 
line & code   & GPU   & CPU \\ \hline
0 & x0 = torch.rand([1024,1024]) & 4 & 0\\
1 & x1 = x0.view(-1,1)   & 4  & 0 \\
2 & y0 = x0.to(`cpu')    & 4  & 4 \\
3 & y1 = x1.to(`cpu')    & 4  & 8 \\
\end{tabular}

% \begin{tablenotes}
% % \item[+] The accuracies for the matched and the mismatched dataset from  the dense model are 84.5 and 84.9 respectively.
% \item[*] Since \textbf{STR} cannot control the sparsity precisely, we report the metrics with our closest achieved sparsity levels, 85.9\% for the 90\% case and 91.8\% for the 94\% case. 
% \end{tablenotes}

\end{threeparttable}
\caption{
LLM fine-tuning may need to use CPU memory to offload large activations.
Lacking cross-device tensor management can lead to redundant copies across devices (especially when the computation graph is complex), which can be particularly undesirable for LLM train-time optimization. For example, although $x0$ and $x1$ are the same tensor with just a different view, when copied to CPU, the resulting tensors $y_0$ and $y_1$ do not share the data storage while $x_0$ and $x_1$ do on GPU.}
\label{cross_device_ex}
\vspace{-0.1 in}
\end{table}

\begin{figure}[!t]
\centering
    \setlength{\tabcolsep}{8 pt}
    \renewcommand{\arraystretch}{0.95}  
   \centering
\mbox{

		\subfigure[Without Marshalling]
		{\includegraphics[width=2.7 in]{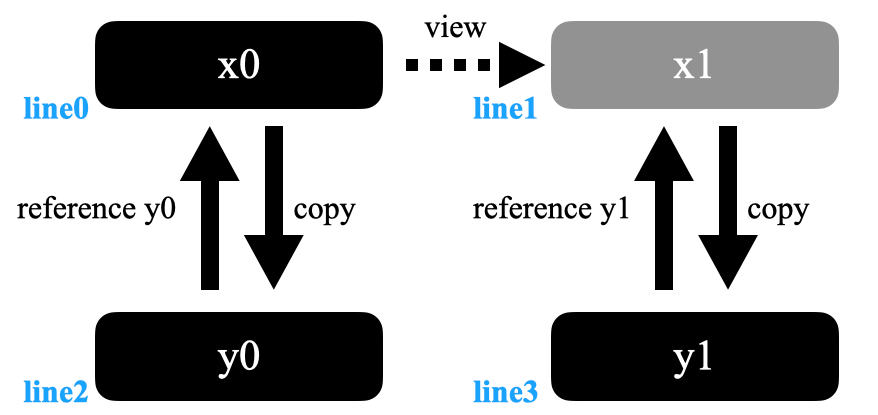}} 
  }
  \mbox{
  \hspace{-0.4 cm}
		\subfigure[With Marshalling ]
		{\includegraphics[width=2.7 in]{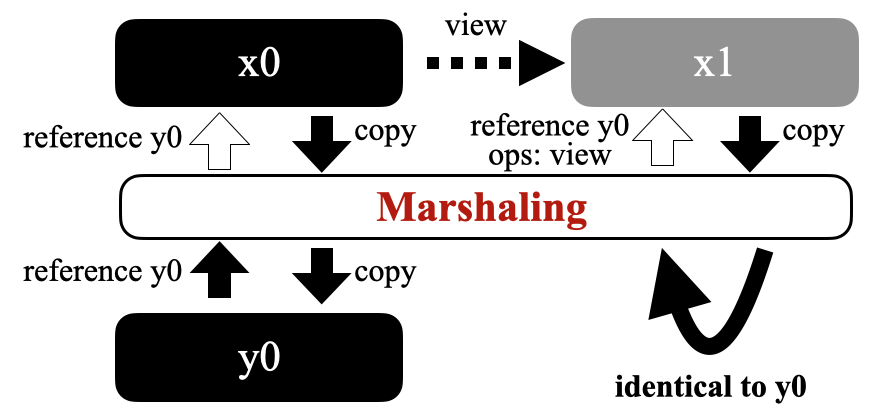}} 
 }

        % \qquad

% \vspace{0.1 in}
\caption{ 
% \mohammad{you need to have time stamp same as the line number in the table on the figure to make the order more clear. And the concept of the ticket has no explenation}
When the proposed cross-device tensor marshalling is applied to the case in Table~\ref{cross_device_ex}, we can avoid duplication on the CPU side, which saves the memory/traffic. Before copying $x_1$ to CPU, our marshaling scheme checks if there exits tensor with the same data storage on the CPU (i.e., $y_0$). If there is, we reuse the reference for $y_0$ along with the required ops (\textit{view} in this case) for future retrieval.}
\label{cross_device_marshall}
    % \vspace{-0.2 in}
\end{figure}

\begin{figure*}[!t]
% \vspace{-0.1 in}

\centering 

\includegraphics[width=6.7in]{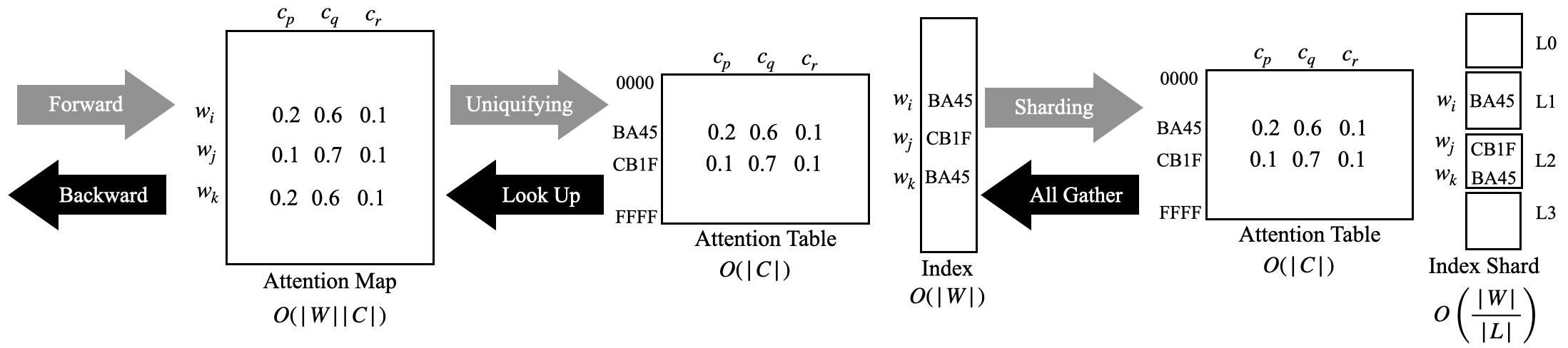}

 \caption{Weight Uniquification and Sharding: since $w_i$ and $w_k$ have the same bit value (BA45), both can share the same attention to centroids in the attention table, yet use the bit value as the offset to the table in the index list.  }
 % \vspace{-0.15 in}
\label{mem_dkm:flow}
\end{figure*}   

To address such inefficiency, we place a marshaling layer as in Fig.~\ref{cross_device_marshall} (b), where the black represents actual data storage and metadata, and the gray indicates only the metadata. Fig.~\ref{cross_device_marshall} (a) illustrates the example in Table~\ref{cross_device_ex} (with the corresponding line numbers) where $x_1$ shares the data layout with $x_0$ but $y_0$ and $y_1$ have independent/duplicated data storage on CPU. By inserting a marshaling layer as in Fig.~\ref{cross_device_marshall} (b), we avoid such redundancy and reduce the GPU-CPU traffic.

We use the \textit{save-tensor-hook} in PyTorch (see~\cite{torch_save_tensor_hook} for reference) 
to implement such a marshaling scheme, where we examine if the same data storage has been already copied. However, checking whether the same tensor exists on the destination device is prohibitively expensive when using a convention scheme like hashing. Therefore, 
when a new tensor   enters our marshaling system, we turn to the forward graph and check if there exists another tensor that is already on CPU and is reachable via only data-storage invariant operations (i.e., \textit{view, transpose}, ...) from the new tensor within a few hops.
If not found, the tensor is copied and a reference to the tensor is generated.
If found, we return the reference of the existing tensor and the list of operations tracing back to the new tensor. For the example in Fig.~\ref{cross_device_marshall} (b), instead of copying $x_1$ to CPU, we simply return the reference to $y_0$ and the \textit{view} operation between $x_1$ and $y_0$.

% Before a tensor is copied to another device, we check if the identical data storage exists on the destination.
% For example, since $x_1$ and $y_0$ have the same data storage, we return the same ticket generated originally for $y_0$, instead of actually copying $x_1$ to CPU as in Fig.~\ref{cross_device_marshall} (b).

% for it.\mohammad{This is not clear at all how you are implementing it.A pseudo code in pytorch here would be very good for clarification} 

Navigating the computation graph costs extra compute cycles, but saving on an unnecessary copy can compensate for such overhead. We found that searching within 4 hops is sufficient to detect all the qualified cases in the computation graph from the original DKM implementation.

\subsection{Weights Uniquification and Sharding}
In most LLM training, 16bit (e.g., BF16 or FP16) is widely used for weights, which means although there are multi-billion parameters in LLMs, there are only  $2^{16}$ unique coefficients due to the bit-width. 
This allows an opportunity to significantly compress the attention map between weights and the centroids, as in Fig~\ref{mem_dkm:flow}. By computing the attention to the centroids once for each unique weight value, the attention map can be converted into an attention table with $O(|C|)$ and the index list with $O(|W|)$.
Note that the number of rows in the attention table is at most 65,536.
% , but in practice, only a few thousand\mohammad{be more specific. How few? 1K-2K? or ~10k? } are needed as most weights are in $(-1,1)$.

% \mohammad{the next two paragraphs have to be rewritten. They are not clear at all}

The index list (denoted $L$) can be further sharded over a set of learners (i.e., GPUs) in a fully synchronous training setup~\cite{fully_sgd}, as the weights are identical in each learner at any moment (thus, attention map and index list too). Such sharding will   bring down the memory complexity to $O(\frac{|W|}{|L|})$.
Uniquifying and sharding come with higher communication and computation costs, as the sharded weights need to be all-gathered and the attention table and index list need to be converted back to the attention map for backward propagation (see Table~\ref{ablation} for the runtime overhead).

% While activations in general cannot be shared due to their data dependency \mohammad{why do we care about activation here?}, the index list can be, as it is not data dependent but weight dependent \mohammad{you have not defined weight vs. data} in a fully synchronous training setup \mohammad{define synchronous training or cite, or be clear why it matters}. Therefore, the index list then can be further sharded over a set of learners \mohammad{what are learners? do you mean different GPUs? or unique values?} (denoted $L$) in the full synchronous training setup, which will additionally  bring down the memory complexity to $O(\frac{|W|}{|L|})$.
% \mohammad{if you have study on the cost in experiment refer to it} 

Assume $\{w_i, w_j, w_k\} \in W$ and $\{c_p, c_q, c_r\} \in C$, which denote the weights and centroids respectively in Fig.~\ref{mem_dkm:flow}. Further consider the case where $\{w_i, w_k \}$ have
the same 16bit representation  $BA45$ and $w_j$ has $CB1F$. Then, when an attention map is computed during forward pass, $w_i$ and $w_k$ shall have the same attention to $C$. 
After uniquification, the attention map is decomposed into an attention table with $O(|C|)$ memory complexity and an index list with $O(|W|)$ complexity. For example, the 16bit value, $BA45$ of $w_i$ and $w_k$ can serve as an offset to the attention table in the index list.
The index list can be further sharded over $|L|$ learners to reduce the complexity in each learner into $O (\frac{|W|}{|L|})$.
The original attention map needs to be reconstructed for backward pass to
stay compatible with the existing autograd implementation. Therefore, we take the reverse steps to restore the attention map by performing all-gather and look-up.  

% $. The attention map from DKM~\cite{dkm}

\begin{table}[!b]
\setlength{\tabcolsep}{4 pt}
\centering
\begin{threeparttable}

\begin{tabular}{c|c|c|ccccc|cccccc} 

\multirow{ 2}{*}{M$^a$ } & \multirow{ 2}{*}{S$^b$ } & \multirow{ 2}{*}{U$^c$ }
       & Memory   & Memmory  & Runtime\\  
&   &  &   (MB) & Reduction ($\times$)  & (sec)\\  \hline
   &   &   &  1600 & 1 & 8.67\\
\checkmark & &  &  544 & 2.9 & 8.97\\
\checkmark &\checkmark  &  &  68 & 23.5&  9.5\\
\checkmark  & &\checkmark  &  97 & 16.4&  15.9\\
\checkmark  & \checkmark &\checkmark  &  12 &129.9&   14.9\\
\end{tabular}

\begin{tablenotes}
% \item[+] The accuracies for the matched and the mismatched dataset from  the dense model are 84.5 and 84.9 respectively.
\item[a] M: using marshaling layer
\item[b] S: using sharding
\item[c] U: using uniquification

\end{tablenotes}

\end{threeparttable}
\caption{Ablation study to understand the effects of each techniques: With the proposed techniques, the memory footprint can be reduced by 130x with 1.7x slow down.}
\label{ablation}
% \vspace{-0.1 in}
\end{table}

\begin{table*}[!t]
\setlength{\tabcolsep}{6 pt}
\centering
\begin{threeparttable}

\begin{tabular}{c|c|c|ccccc|cccccc} 
\multirow{2}{*}{Method}  & \multirow{2}{*}{bits} & Model&     \multicolumn{5}{c|}{Common Sense Reasoning} &    \multicolumn{2}{c}{Few-shot} \\\cline{4-10}
                          &       &     Size(GB)     &  PIQA & HellaSwag & Winograde &  ARC-e  & ARC-c  & TriviaQA  & MMLU\\  \hline
LLaMA-7B      & 16   & 12.6  & 79.3  & 76.1  & 70.0  & 73.0  & 48.0  & 57.0  & 35.2 \\
RTN	          & 4	& 3.5	& 77.3	& 72.7	& 66.9	& 68.8	& 46.4	& 44.9	& 28.9 \\
% SmoothQuant	  & 4	& 3.5	& 76.4	& 69.4	& 66.8  & 66.9  &	43.0& 40.0	& 29.0 \\
% GPTQ     	  & 4	& 3.7	& 76.0	& 69.5	& 66.7	& 66.9	&		& & \\
GPTQ g128$^{c}$	  & 4	& 3.7	& 77.2	& 54.0	& 65.7	& 61.6	&	--$^{a}$	 & --& --\\
AWQ g128	  & 4	& 3.7	& 78.1	& 55.8	& 65.8	& 66.8	&	--	 & --& --\\
LLM-QAT	      & 4	& 3.5	& 78.3	& 74.0	& 69.0	& 70.0	& 45.0	&	50.8	& 30.8 \\
GPTQ g128	  & 3	& 3.0	& 70.9	& 46.8	& 60.9	& 66.1	&	--	& --& --\\
AWQ g128	  & 3	& 3.0	& 76.7	& 53.6	& 66.1	& 65.7	&	--	& --&-- \\
\prjname 	  & 3	& 2.5	& 77.7	& 54.6	& 66.1	& 72.3	& 40.3	&	35.2$^{b}$	& 30.3 \\

\end{tabular}

\begin{tablenotes}
% \item[+] The accuracies for the matched and the mismatched dataset from  the dense model are 84.5 and 84.9 respectively.
\item[]  $^a$ The result is not reported for the corresponding scheme; $^b$ One-shot is applied; $^c$ Group size is 128.
\end{tablenotes}

\end{threeparttable}
\caption{When compared our techniques against the state-of-the-art compression scheme, \prjname offered the smallest model size, yet similar or  better accuracy for the broader set of benchmarks with the 3bit compressed LLaMA 7B model.}
\label{result_llm_accuracy}
% \vspace{-0.1 in}
\end{table*}

\section{Experimental Results}
We used the PyTorch 2.0.01 and applied Fully Sharded Data Parallel (FSDP) to fine-tune the pretrained LLaMA 7B model in brainfloat16 with the Alpaca dataset~\cite{alpaca}. 
We fine-tuned for 2 epochs while compressing the model on a single node with $8\times$ A100-80GB GPUs using  \prjname. The maximum sequence length during fine-tuning was 256.
We used AdamW optimizer with learning rate as 5e-5, weight decay as 0, and betas as $(0.9, 0.95)$. The global batch size is 64, and the gradient norm clipping with 1.0 is used.

% \mohammad{why do you start with Ablation? first show the result of the main experiment and then do the ablation}

\subsection{LLM Accuracy}
We compared \prjname against other quantization-based compression schemes: round-to-nearest (RTN), SmoothQuant, GPTQ~\cite{frantar2023gptq}, AWQ~\cite{lin2023awq} and LLM-QAT~\cite{liu2023llmqat}. For \prjname, we also compressed the embedding layers with 8 bits.

% \mohammad{be clear on your training set-up . What was the tuning set? what was the optimizer,  learning rate or any other parameters ? How long did you finetune? What waas the value of $L$ ?}

Table~\ref{result_llm_accuracy} reports the accuracy with Common Sense Reasoning, and Few-Shot benchmarks with the compressed LLaMA 7B models from each technique.  

\begin{itemize}
    \item \prjname allows the 3bit compressed LLaMA 7B model to outperform all other schemes in the 3bit configuration.
    \item  \prjname even delivers the best accuracy for ARC-e benchmarks across 3 and 4bit configurations.
    \item  \prjname yields the competitive performance for PIQA and MMLU benchmarks with 4bit compressed models.
\end{itemize}

\subsection{Ablation Study}
For the ablation study, we made an example with one attention layer from the LLaMA 7B decoder stack and measured the trade-off between the memory footprint vs. the forward-backward speed with  3bit compression, as shown in Table~\ref{ablation}.

Cross-device tensor marshaling alone reduces the memory footprint by 2.9$\times$ with little runtime overhead, and the additional savings of 23.5$\times$ and 16.4$\times$ are achieved with sharding and uniquification, respectively. When all techniques combined, as in Fig.~\ref{mem_dkm:flow}, \prjname offered about 130x reduction.
Although these steps require extra computation/communications (i.e., all-gather), the runtime overhead is insignificant, as the traffic between GPU and CPU has decreased substantially.

\section{Conclusion}
In this work, we propose a memory-efficient differentiable weight clustering scheme, \prjname, to provide train-time compression for LLMs. With the proposed techniques, the memory consumption was reduced by almost 130x, and the resulting 3bit compressed LLaMA model yields state-of-the-art accuracy on various LLM-harness benchmarks.

% \section{Acknowledgement}
% We sine

%%%%%%% -- PAPER CONTENT ENDS -- %%%%%%%%

%%%%%%%%% -- BIB STYLE AND FILE -- %%%%%%%%
\bibliographystyle{IEEEtranS}
\bibliography{main}

% Generated by IEEEtranS.bst, version: 1.13 (2008/09/30)
\begin{thebibliography}{10}
\providecommand{\url}[1]{#1}
\csname url@samestyle\endcsname
\providecommand{\newblock}{\relax}
\providecommand{\bibinfo}[2]{#2}
\providecommand{\BIBentrySTDinterwordspacing}{\spaceskip=0pt\relax}
\providecommand{\BIBentryALTinterwordstretchfactor}{4}
\providecommand{\BIBentryALTinterwordspacing}{\spaceskip=\fontdimen2\font plus
\BIBentryALTinterwordstretchfactor\fontdimen3\font minus
  \fontdimen4\font\relax}
\providecommand{\BIBforeignlanguage}[2]{{%
\expandafter\ifx\csname l@#1\endcsname\relax
\typeout{** WARNING: IEEEtranS.bst: No hyphenation pattern has been}%
\typeout{** loaded for the language `#1'. Using the pattern for}%
\typeout{** the default language instead.}%
\else
\language=\csname l@#1\endcsname
\fi
#2}}
\providecommand{\BIBdecl}{\relax}
\BIBdecl

\bibitem{coreml}
\url{https://coremltools.readme.io/docs/training-time-palettization}.

\bibitem{torch_save_tensor_hook}
\url{https://pytorch.org/docs/stable/autograd.html#torch.autograd.graph.saved_tensors_hooks}.

\bibitem{dkm}
M.~Cho, K.~Alizadeh{-}Vahid, S.~Adya, and M.~Rastegari, ``{{DKM:}
  Differentiable K-Means Clustering Layer for Neural Network Compression},'' in
  \emph{International Conference on Learning Representations}, 2022.

\bibitem{fully_sgd}
J.~Dean, G.~Corrado, R.~Monga, K.~Chen, M.~Devin, M.~Mao, M.~a. Ranzato,
  A.~Senior, P.~Tucker, K.~Yang, Q.~Le, and A.~Ng, ``Large scale distributed
  deep networks,'' in \emph{Advances in Neural Information Processing Systems},
  2012.

\bibitem{hawq-v2}
Z.~Dong, Z.~Yao, D.~Arfeen, A.~Gholami, M.~W. Mahoney, and K.~Keutzer,
  ``{HAWQ-V2: Hessian Aware trace-Weighted Quantization of Neural Networks},''
  in \emph{Advances in Neural Information Processing Systems}, 2020.

\bibitem{quant_noise}
A.~Fan, P.~Stock, B.~Graham, E.~Grave, R.~Gribonval, H.~J{\'{e}}gou, and
  A.~Joulin, ``Training with quantization noise for extreme model
  compression,'' in \emph{International Conference on Learning
  Representations}, 2021.

\bibitem{frantar2023gptq}
E.~Frantar, S.~Ashkboos, T.~Hoefler, and D.~Alistarh, ``{GPTQ: Accurate
  Post-Training Quantization for Generative Pre-trained Transformers},'' in
  \emph{arXiv}, 2023.

\bibitem{deepcomp_iclr16}
S.~Han, H.~Mao, and W.~J. Dally, ``Deep compression: Compressing deep neural
  network with pruning, trained quantization and huffman coding,'' in
  \emph{International Conference on Learning Representations}, 2016.

\bibitem{lin2023awq}
J.~Lin, J.~Tang, H.~Tang, S.~Yang, X.~Dang, and S.~Han, ``{AWQ:
  Activation-aware Weight Quantization for LLM Compression and Acceleration},''
  \emph{arXiv}, 2023.

\bibitem{liu2023llmqat}
Z.~Liu, B.~Oguz, C.~Zhao, E.~Chang, P.~Stock, Y.~Mehdad, Y.~Shi,
  R.~Krishnamoorthi, and V.~Chandra, ``{LLM-QAT: Data-Free Quantization Aware
  Training for Large Language Models},'' \emph{arXiv}, 2023.

\bibitem{ouyang2022training}
L.~Ouyang, J.~Wu, X.~Jiang, D.~Almeida, C.~L. Wainwright, P.~Mishkin, C.~Zhang,
  S.~Agarwal, K.~Slama, A.~Ray, J.~Schulman, J.~Hilton, F.~Kelton, L.~Miller,
  M.~Simens, A.~Askell, P.~Welinder, P.~Christiano, J.~Leike, and R.~Lowe,
  ``Training language models to follow instructions with human feedback,'' in
  \emph{Advances in Neural Information Processing Systems}, 2022.

\bibitem{Park_2018_ECCV}
E.~Park, S.~Yoo, and P.~Vajda, ``Value-aware quantization for training and
  inference of neural networks,'' in \emph{European Conference on Computer
  Vision}, 2018.

\bibitem{park2019lookahead}
S.~Park, J.~Lee, S.~Mo, and J.~Shin, ``Lookahead: A far-sighted alternative of
  magnitude-based pruning,'' in \emph{International Conference on Learning
  Representations}, 2019.

\bibitem{polino2018model}
A.~Polino, R.~Pascanu, and D.-A. Alistarh, ``Model compression via distillation
  and quantization,'' in \emph{International Conference on Learning
  Representations}, 2018.

\bibitem{xor_net}
M.~Rastegari, V.~Ordonez, J.~Redmon, and A.~Farhadi, ``Xnor-net: Imagenet
  classification using binary convolutional neural networks,'' in
  \emph{European Conference on Computer Vision}.\hskip 1em plus 0.5em minus
  0.4em\relax Springer, 2016, pp. 525--542.

\bibitem{stock2020bit}
P.~Stock, A.~Joulin, R.~Gribonval, B.~Graham, and H.~Jégou, ``And the bit goes
  down: Revisiting the quantization of neural networks,'' in
  \emph{International Conference on Learning Representations}, 2020.

\bibitem{alpaca}
R.~Taori, I.~Gulrajani, T.~Zhang, Y.~Dubois, X.~Li, C.~Guestrin, P.~Liang, and
  T.~B. Hashimoto, ``{Stanford Alpaca: An Instruction-following LLaMA model},''
  \url{https://github.com/tatsu-lab/stanford_alpaca}, 2023.

\bibitem{touvron2023LLaMA}
H.~Touvron, T.~Lavril, G.~Izacard, X.~Martinet, M.-A. Lachaux, T.~Lacroix,
  B.~Rozière, N.~Goyal, E.~Hambro, F.~Azhar, A.~Rodriguez, A.~Joulin,
  E.~Grave, and G.~Lample, ``Llama: Open and efficient foundation language
  models,'' in \emph{arXiv}, 2023.

\bibitem{ullrich2017soft}
K.~Ullrich, E.~Meeds, and M.~Welling, ``Soft weight-sharing for neural network
  compression,'' in \emph{International Conference on Learning
  Representations}, 2017.

\bibitem{haq}
K.~Wang, Z.~Liu, Y.~Lin, J.~Lin, and S.~Han, ``Haq: Hardware-aware automated
  quantization with mixed precision,'' in \emph{Proceedings of the IEEE
  Conference on Computer Vision and Pattern Recognition}, 2019.

\bibitem{pmlr-v80-wu18h}
J.~Wu, Y.~Wang, Z.~Wu, Z.~Wang, A.~Veeraraghavan, and Y.~Lin, ``Deep k-means:
  Re-training and parameter sharing with harder cluster assignments for
  compressing deep convolutions,'' in \emph{International Conference on Machine
  Learning}, 2018.

\bibitem{yu2018nisp}
R.~Yu, A.~Li, C.-F. Chen, J.-H. Lai, V.~I. Morariu, X.~Han, M.~Gao, C.-Y. Lin,
  and L.~S. Davis, ``Nisp: Pruning networks using neuron importance score
  propagation,'' in \emph{Proceedings of the IEEE Conference on Computer Vision
  and Pattern Recognition}, 2018, pp. 9194--9203.

\bibitem{zhang2022opt}
S.~Zhang, S.~Roller, N.~Goyal, M.~Artetxe, M.~Chen, S.~Chen, C.~Dewan, M.~Diab,
  X.~Li, X.~V. Lin, T.~Mihaylov, M.~Ott, S.~Shleifer, K.~Shuster, D.~Simig,
  P.~S. Koura, A.~Sridhar, T.~Wang, and L.~Zettlemoyer, ``Opt: Open pre-trained
  transformer language models,'' in \emph{arXiv}, 2022.

\bibitem{zhou2019neural}
D.~Zhou, X.~Jin, Q.~Hou, K.~Wang, J.~Yang, and J.~Feng, ``Neural epitome search
  for architecture-agnostic network compression,'' in \emph{International
  Conference on Learning Representations}, 2019.

\end{thebibliography}
%%%%%%%%%%%%%%%%%%%%%%%%%%%%%%%%%%%%

\end{document}